\documentclass[final,1p,times]{elsarticle}

\usepackage{amssymb}
\usepackage{comment}
\usepackage{graphicx}
\usepackage{physics}
\usepackage{svg}
\usepackage{url}
\usepackage{soul}

\newcommand{\avgk}{\langle k \rangle}
\newcommand{\roll}{\textit{swiss roll}}
\newcommand{\scurve}{\textit{s curve}}
\newcommand{\nclasses}{\texttt{n\_classes}}
\newcommand{\nreps}{\texttt{n\_reps}}

\journal{Neural Networks}

\begin{document}

\begin{frontmatter}



\title{Beyond Multilayer Perceptrons: Investigating Complex Topologies in Neural Networks}


\author[tv]{Tommaso Boccato\corref{cor}}
\ead{tommaso.boccato@uniroma2.it}
\author[tv]{Matteo Ferrante}
\ead{matteo.ferrante@uniroma2.it}
\author[tv]{Andrea Duggento\fnref{ec}}
\ead{duggento@med.uniroma2.it}
\author[tv,mc]{Nicola Toschi\fnref{ec}}
\ead{toschi@med.uniroma2.it}

\affiliation[tv]{organization={Department of Biomedicine and Prevention, University of Rome Tor Vergata},
            city={Rome},
            country={Italy}}
\affiliation[mc]{organization={A.A. Martinos Center for Biomedical Imaging and Harvard Medical School},
            city={Boston},
            country={USA}}

\cortext[cor]{Corresponding author}
\fntext[ec]{These authors contributed equally to this work}

\begin{abstract}
In this study, we explore the impact of network topology on the approximation capabilities of artificial neural networks (ANNs), with a particular focus on complex topologies. We propose a novel methodology for constructing complex ANNs based on various topologies, including Barabási-Albert, Erdős-Rényi, Watts-Strogatz, and multilayer perceptrons (MLPs). The constructed networks are evaluated on synthetic datasets generated from manifold learning generators, with varying levels of task difficulty and noise, \textcolor{black}{and on real-world datasets from the UCI suite}.

Our findings reveal that complex topologies lead to superior performance in high-difficulty regimes compared to traditional MLPs. This performance advantage is attributed to the ability of complex networks to exploit the compositionality of the underlying target function. However, this benefit comes at the cost of increased forward-pass computation time and reduced robustness to graph damage.

Additionally, we investigate the relationship between various topological attributes and model performance. Our analysis shows that no single attribute can account for the observed performance differences, suggesting that the influence of network topology on approximation capabilities may be more intricate than a simple correlation with individual topological attributes.

Our study sheds light on the potential of complex topologies for enhancing the performance of ANNs and provides a foundation for future research exploring the interplay between multiple topological attributes and their impact on model performance.

\end{abstract}




\end{frontmatter}


\section{Introduction}

\begin{figure}[h]
\centering
\includegraphics[width=.9\textwidth]{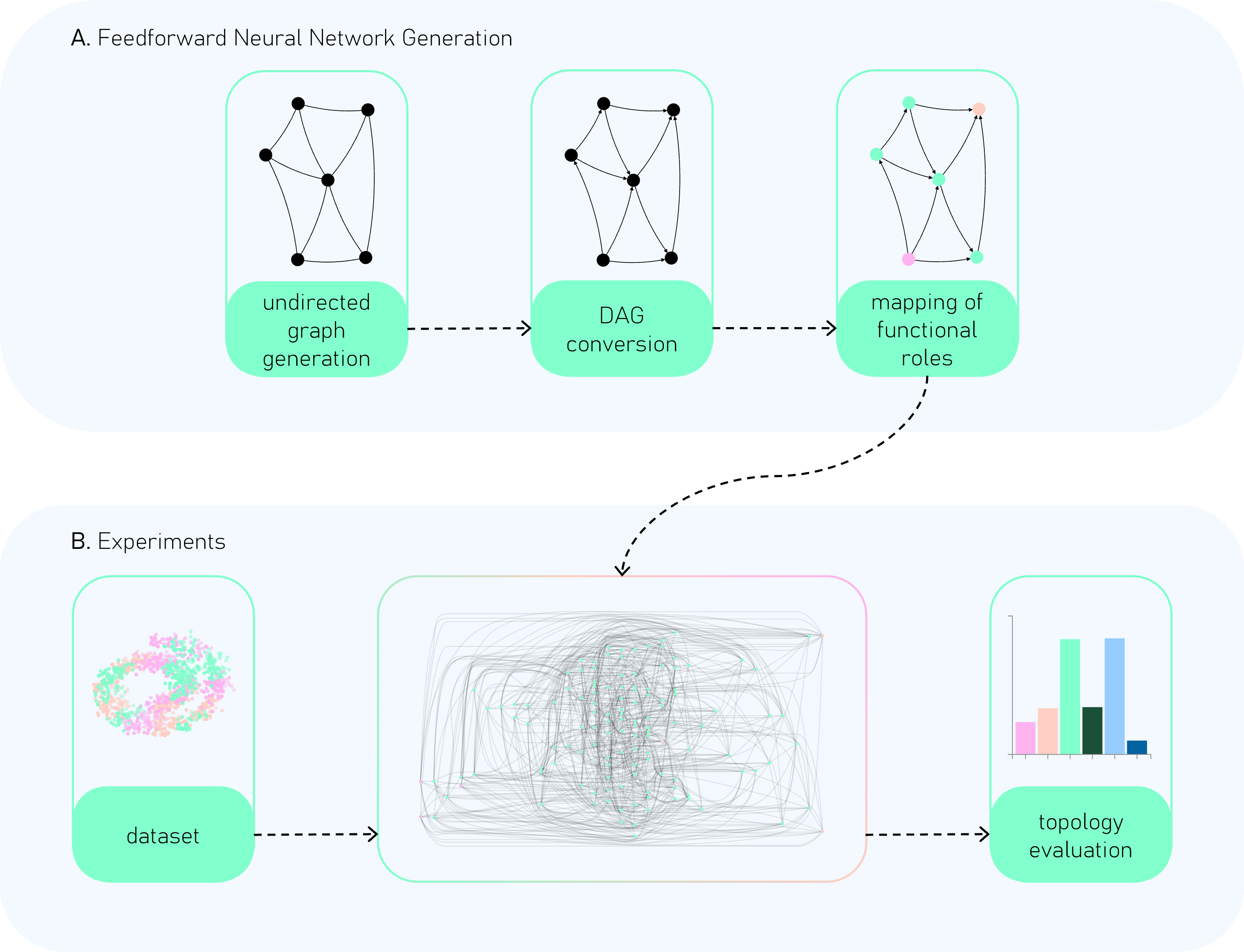}
\caption{Overview of the topology exploration process. \textbf{Top}: Feedforward neural network (NN) generation. All studied models are constructed using the same three-step procedure. First, we generate an undirected graph with a predetermined degree distribution. Then, we set edge directions and map computational operations to the network nodes. \textbf{Bottom}: Experiments. For each investigated topology, we sample multiple graphs from the same degree distribution. The corresponding NNs are trained on one of the benchmark datasets. The resulting test accuracies are collected and stored for subsequent analyses.\label{fig:overview}}
\end{figure}

Modern neural architectures are widely believed to draw significant design inspiration from biological neuronal networks. The artificial neuron, the fundamental functional unit of neural networks (NNs), is based on the McCulloch-Pitts unit \cite{fitch_1944}, sharing conceptual similarities with its biological counterpart. Additionally, state-of-the-art convolutional NNs incorporate several operations directly inspired by the mammalian primary visual cortex, such as nonlinear transduction, divisive normalization, and maximum-based pooling of inputs. However, these architectures may be among the few examples where the evolutionary structural and functional properties of neuronal systems have been genuinely relevant for NN design. Indeed, the topology of biological connectomes has not yet been translated into deep learning model engineering.

Due to the ease of implementation and deployment, widely-used neural architectures predominantly feature a regular structure resembling a sequence of functional blocks (e.g., neuronal layers). The underlying multipartite graph of a multilayer perceptron (MLP) is typically controlled by a few hyperparameters that define its basic topological properties: depth, width, and layer sizes. Only recently have computer vision engineers transitioned from chain-like structures \cite{2014arXiv1409.1556S} to more elaborate connectivity patterns \cite{7780459,8099726,Xie_2019_ICCV} (e.g., skip connections, complete graphs). \textcolor{black}{Nevertheless, biological neuronal networks display much richer and less templated wirings at both the micro- and macro-scale \cite{FORNITO2013426,osti_10289849,osti_10351662,osti_10351663}. For example, considering synaptic connections between individual neurons, the \textit{C. elegans} nematode features a hierarchical modular \cite{bassett2010efficient} connectome, wherein hubs with high betweenness centrality are efficiently interconnected \cite{barthelemy2004betweenness,towlson2013rich}. Moreover, the strength distribution of the adult Drosophila central brain closely follows a power law with an exponential cutoff \cite{scheffer_connectome_2020}.}

As a result, the relationship between the graph structure of a NN and its predictive abilities remains unclear. In the literature, there is evidence that complex networks can be advantageous in terms of predictive accuracy and parameter efficiency \cite{KAVIANI2021115073}. However, past attempts to investigate this connection have yielded conflicting results that are difficult to generalize outside the investigated context. The first experiment on complex NNs was performed in 2005 by Simard et al., who trained a randomly rewired MLP on random binary patterns \cite{simard_fastest_2005}. Nearly a decade later, Erkaymaz and his collaborators employed the same experimental setup on various real-life problems \cite{erkaymaz_performance_2012,erkaymaz_impact_nodate,erkaymaz_impact_2016,erkaymaz_performance_2017} (e.g., diabetes diagnosis, performance prediction of solar air collectors). The best-performing models featured a number of rewirings consistent with the small-world regime. However, all assessed topologies were constrained by MLP-random interpolation. In \cite{basili_evolving_2007}, an MLP and a NN generated following the Barabási-Albert (BA) procedure were compared on a chemical process modeling problem. Both models were trained with an evolutionary algorithm, but the MLP achieved a lower RMSE. The \textit{learning matrix} \cite{monteiro_model_2016}, a sequential algorithm for the forward/backward pass of arbitrary directed acyclic graphs (DAGs), enabled the evaluation of several well-known complex networks on classification \cite{monteiro_model_2016} and regression \cite{platt_computational_2019} tasks. The experiments included random and small-world networks, two topologies based on ``preferential attachment'', a complete graph, and a \textit{C. elegans} subnetwork \cite{DBLP:journals/jcns/DunnLPC04}. Nevertheless, the learning matrix's time complexity limited the network sizes (i.e., 26 nodes), and for each task, a different winning topology emerged, including the MLP. Also Stier et al. successfully trained BA- and WS-based (Watts-Strogatz) NNs with backpropagation \cite{STIER2019107} on the MNIST classification task \cite{726791} by placing the generated networks between two fully-connected layers. While this design choice was made in order to adapt the architecture to the dimensionalities of the input/output, it may represent a confound when disentangling the contributions of the different network modules to the overall classification performance. Some recent works have instead focused on multipartite sparse graphs \cite{mocanu_scalable_2018,you_graph_2020}. While these architectures outperformed the complete baselines, their topological complexity was entirely encoded within the connections between adjacent layers. \textcolor{black}{Another area of research that explores NNs characterized by complex graphs is the Lottery Ticket Hypothesis (LTH) \cite{DBLP:journals/corr/abs-1803-03635}. The LTH posits that deep NNs contain subnetworks, often referred to as ``winning tickets'', with optimized initial weights. When trained in isolation, these subnetworks can achieve high performance on specific tasks. However, it is important to note that these subnetworks are restricted to the ``mother'' architectures, which typically consist of multipartite graphs or chain-like macro-scale networks.}

\begin{figure}
\centering
\includegraphics[width=.9\textwidth]{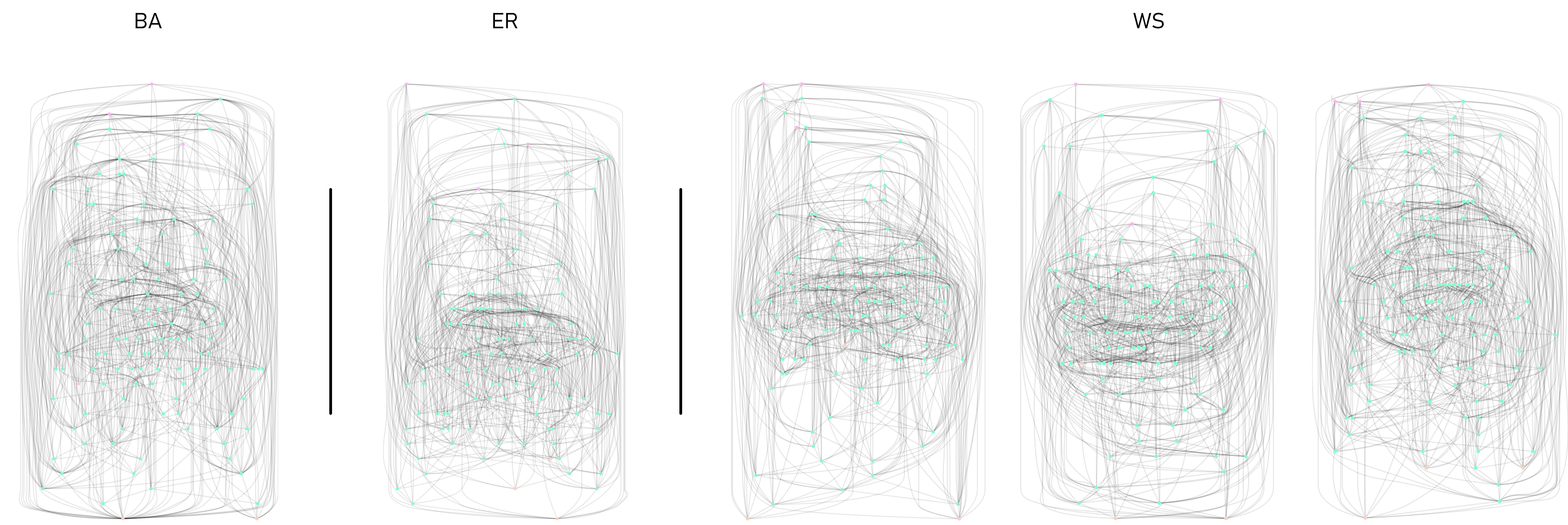}
\caption{Example feedforward NNs (128 neurons, 732 synaptic connections) based on complex topologies: scale-free (BA), random (ER), and small-world (WS). All graphs are directed and acyclic. Information flows from top to bottom. Input, hidden, and output units are denoted in pink, green, and orange, respectively. Since the networks are defined at the micro-scale, hidden and output nodes implement weighted sums over the incoming edges. In the hidden units, the computational operation is followed by an activation function. The activations of nodes located on the same horizontal layer can be computed in parallel.\label{fig:complex-nets}}
\end{figure}

We propose the hypothesis that, given the same number of nodes (i.e., neurons) and edges (i.e., parameters), a complex NN might exhibit superior predictive abilities compared to classical, more regularly structured MLPs. Unlike previous studies, we conduct a systematic exploration (of which we have reported an overview in Figure \ref{fig:overview}) of random, scale-free, and small-world graphs (Figure \ref{fig:complex-nets}) on synthetic classification tasks\footnote{The source code for our experiments will soon be available at \url{https://github.com/BoCtrl-C}.}, with particular emphasis on the following:
\begin{itemize}
  \item \textbf{Network size.} The defining properties of a complex topology often emerge in large-scale networks. For example, the second moment of a power-law degree distribution diverges only in the $N \rightarrow \infty$ limit \cite{barabasi2016network}, where $N$ is the network size\footnote{The proposition holds when the degree exponent is smaller than 3.}. The networks in \cite{monteiro_model_2016,platt_computational_2019} have 15 and 26 nodes, respectively. We trained models with 128 neurons.
\item \textbf{Dataset size.} The \textit{estimation error} achieved by a predictor depends on the training set size: the greater the number of samples, the lower the error \cite{shalev-shwartz_ben-david_2014}. Except for studies based on multipartite graphs, all previous research works in a small-data regime. Our datasets are three times larger than those used before.
\item \textbf{Hyperparameter optimization.} Learning rate and batch size are crucial in minimizing the loss function. Ref. \cite{monteiro_model_2016} is the only one that considers finding the optimal learning rate. The role of batch size has never been investigated. Each DAG, however, could be characterized by its optimal combination of hyperparameters. Hence, we optimized the learning rate and batch size for each topology.
\end{itemize}
\color{black}
Further, we present a series of supplementary experiments aimed at exploring the suitability of non-standard topologies and the applicability of our results to real-world data.
\color{black}

\color{black}
\section{Theory}\label{sec:theory}

In this section, we briefly report on the network science theory behind graph generators. These graph models are involved in generating the NNs employed in our investigation, as discussed in Section \ref{sec:methods}.\smallskip

\noindent\textbf{Erdős-Rényi (ER).} An ER graph \cite{erdHos1960evolution}, or \textit{random network}, is uniformly sampled from the set of all graphs with $N$ nodes and $L$ edges. For $N \gg \avgk$, the degree distribution of a random graph is well approximated by a Poisson distribution: $p_k = e^{-\avgk}\frac{\avgk^k}{k!}$; $k$ and $\avgk$ represent node degree and average degree, respectively.\smallskip

\noindent\textbf{Watts-Strogatz (WS).} The WS generator \cite{watts1998collective} aims to create graphs that exhibit both high clustering and the \textit{small-world} property; this is achieved by interpolating \textit{lattices} with random networks. The generation starts from a ring in which nodes are connected to their immediate neighbors, and the links are then randomly rewired with probability $p$.\smallskip

\noindent\textbf{Barabási-Albert (BA).} The well-known BA model \cite{albert2002statistical} can be used to generate networks characterized by the $p_k \propto k^{-3}$ \textit{scale-free} degree distribution. Given that the model is inspired by the growth of real networks, the generative procedure iteratively attaches nodes with $m$ stubs to a graph that evolves from an initial star of $m + 1$ nodes. Node additions respond to the preferential attachment mechanism: the probability that a stub reaches a node is proportional to the degree of the latter.\smallskip

\noindent\textbf{Multilayer Perceptron (MLP).} The networks underlying MLPs are called multipartite graphs. In a multipartite graph (i.e., a sequence of bipartite graphs) nodes are partitioned into layers, and each layer can only be connected with the adjacent ones; no intra-layer link is allowed. Additionally, inter-layer connections have to form \textit{bicliques} (i.e., fully-connected bipartite graphs).\smallskip

We have reported a comprehensive description of the Stochastic Block Model (SBM) in \ref{sec:sbm}.
\color{black}

\section{Methods}\label{sec:methods}

The following sections present our methodology. Section \ref{sec:datasets} describes how our benchmark datasets are constructed. In Section \ref{sec:feedforward}, we provide details on the proposed NN generation pipeline. Finally, Section \ref{sec:experiments} details out the experimental protocol.

\subsection{Datasets}\label{sec:datasets}

\begin{figure}
\centering
\includegraphics[width=.75\textwidth]{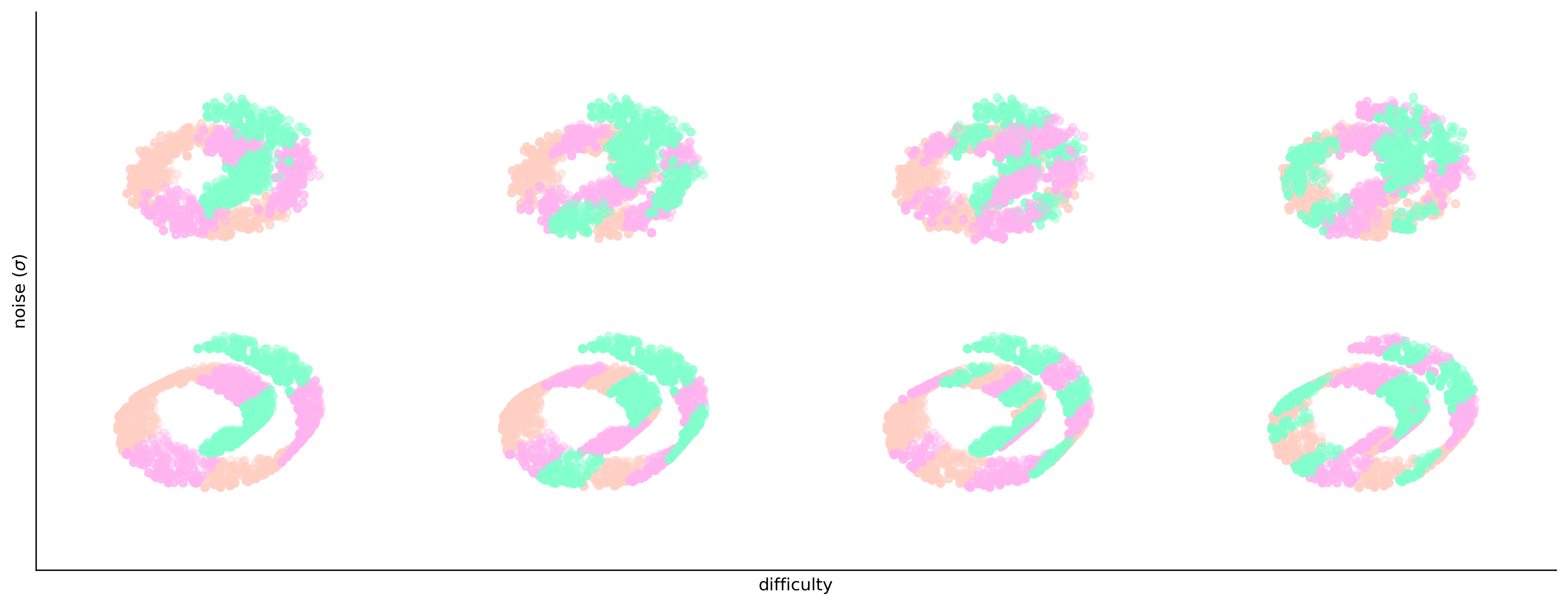}
\includegraphics[width=.75\textwidth]{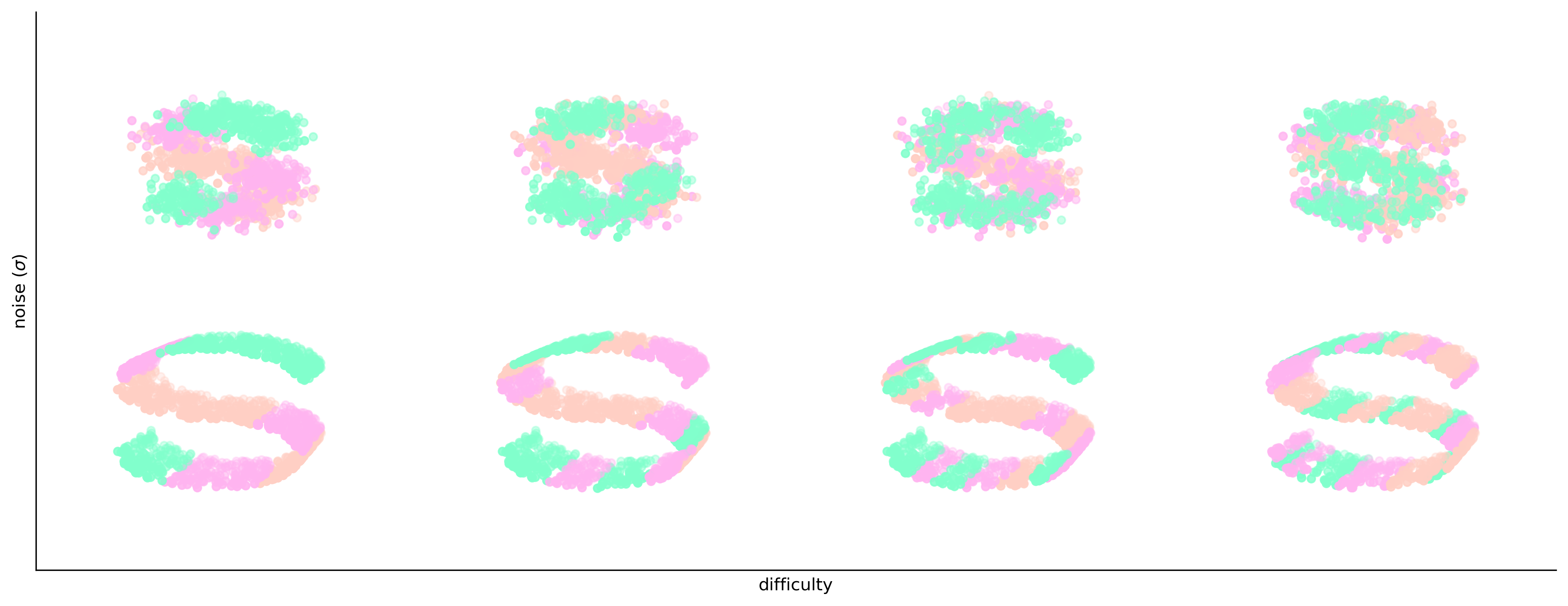}
\caption{Benchmark classification datasets. \textbf{Top}: the \roll{}. \textbf{Bottom}: the \scurve{}. Each dataset is composed of 3D points divided into multiple segments. Classes are color-coded. Datasets differ in terms of difficulty ($x$ axis) and noise ($y$ axis). \label{fig:datasets}}
\end{figure}

The foundation of the datasets developed, as displayed in Figure \ref{fig:datasets}, is established by manifold learning generators\footnote{\url{https://scikit-learn.org/stable/datasets/sample_generators.html}} provided by the \texttt{scikit-learn} machine learning (ML) library \cite{scikit-learn}. To modify the generators for classification purposes, 3D points sampled from one of the available curves (\scurve{} and \roll{}) are segmented into $\nclasses{} \times \nreps{}$ portions based on their univariate position relative to the primary dimension of the manifold samples. As the term implies, \nclasses{} refers to the number of classes involved in the considered classification. \textcolor{black}{Each segment is then arbitrarily allocated to a class, maintaining task balance (i.e., precisely \nreps{} segments have the same label). We define \nreps{} as the task \textit{difficulty}.} An additional aspect of our datasets is the standard deviation $\sigma$ of the Gaussian noise that can be added to the points. The generation procedure is finalized with a min-max normalization.

\subsection{Feedforward Neural Networks}\label{sec:feedforward}
\textcolor{black}{All trainable models are produced following the same 3-step procedure and share $N$ and $L$.} Consequently, NNs exhibit identical density and parameter counts.\smallskip

\noindent\textbf{Undirected Graph Generation.} The initial step in creating a NN involves sampling an undirected graph using the generators detailed in Section \ref{sec:theory}. Once $N$ and $L$ are established, all models exhibit a single parameter configuration compatible with the required density\footnote{This statement is accurate if the number of MLP layers is predetermined.}. The WS generator is the sole exception: the probability $p$ is allowed to vary between 0 and 1. If the generator is limited to sample networks with a number of links from a finite set (e.g., $L = m + (N - m - 1)m$ according to the BA model), we first generate a graph with slightly higher density than the target before randomly eliminating excess edges. After obtaining the graph, we confirm the existence of a single connected component.\smallskip

\color{black}
\noindent\textbf{Directed Acyclic Graph (DAG) Conversion.} Before performing any calculations, the direction for information propagation through the network links must be determined; this is accomplished by randomly assigning, without replacement, an integer index from $\{1, \dots, N\}$ to the network nodes. It can be shown that the directed graph obtained by setting the direction of each edge from the node with a lower index to the node with a higher index is free of cycles \cite{bondy1976graph}. However, this conversion results in an unpredictable number of sources and sinks. Since classification tasks typically involve a pre-defined number of input features and output classes, it is necessary to resolve such network-task discrepancies. To address this issue, we developed a straightforward heuristic capable of adjusting DAGs without altering the underlying undirected graphs.\smallskip
\color{black}

\noindent\textbf{Mapping of Functional Roles.} The last step of the presented procedure consists in mapping computational operations to the DAG nodes. Working at the micro-scale (i.e., connections between single neurons), the operations allowed are two. Source nodes implement constant functions; their role, indeed, is to feed the network with the initial conditions for computations. Hidden and sink nodes, instead, perform a weighted sum over the incoming edges, followed by an activation function:
\begin{equation}
    a_v = \sigma\Bigg(\sum_u w_{uv}a_u + b\Bigg)
\end{equation}
where $a_v$ is the activation of node $v$, $\sigma$ denotes the activation function\footnote{Depending on the context, we use the same $\sigma$ notation for both the standard deviation of the dataset noise and the activation function.} (SELU \cite{NIPS2017_5d44ee6f} for hidden nodes and the identity function for sinks), $u$ represents the generic predecessor of $v$, $w_{uv}$ is the weight associated with edge $(u, v)$ and $b$ the bias. In order to implement the map of functional roles, we made use of the 4Ward library\footnote{\url{https://github.com/BoCtrl-C/forward}} \cite{https://doi.org/10.48550/arxiv.2209.02037}, developed for the purpose. Starting from a DAG, the package returns a working NN deployable as a PyTorch \texttt{Module}.

\subsection{Experiments}\label{sec:experiments}

In this section we report on the experimental protocol designed to compare the performance of the various graph topologies investigated.\smallskip

\noindent\textbf{Dataset Partitioning.} Each generated dataset is randomly divided into 3 non-overlapping subsets: the train, validation and test splits. All model trainings are performed over the train split while the validation split is exploited in validation epochs and hyperparameter optimization. Test samples, instead, are accessed only in the evaluation of the final models.\smallskip

\noindent\textbf{Model Training.} Models are trained by minimizing cross entropy with the Adam \cite{DBLP:journals/corr/KingmaB14} optimizer ($\beta_1 = 0.9$, $\beta_2 = 0.999$). A scheduler reduces the learning rate by a factor of 0.5 if no improvement is seen on the validation loss for 10 epochs. The training procedure ends when learning stagnates (w.r.t. the validation loss) for 15 epochs, and the model weights corresponding to the epoch in which the minimum validation loss has been achieved are saved.\smallskip

\noindent\textbf{Hyparameter Optimization.} Hyperparameters are optimized through a grid search over a predefined 2D space (i.e., learning rate/batch size). We generate networks of the same topological family starting from 5 different random seeds. In the MLP case, models differ only in the weight initialization. For each parameter pair, the 5 models are trained accordingly, and the resulting best validation losses are collected. Then, the learning rate and batch size that minimize the median validation loss computed across the generation seeds are selected as the optimal hyperparameters of the considered graph family.\smallskip

\noindent\textbf{Topology Evaluation.} Once the optimal learning rate and batch size are found, we train 15 new models characterized by the considered topology and compute mean classification accuracy and standard deviation on the dataset test split. The procedure is repeated for each investigated graph family and a Kruskal-Wallis (H-test) \cite{10.2307/2280779} is performed in order to test the null hypothesis that the medians of all accuracy populations are equal. If the null hypothesis is rejected, a Mann-Whitney (U-test) \cite{10.2307/2236101} post hoc analysis follows.\smallskip

\noindent\textbf{Robustness Analysis.} We use the final trained models in a \textit{graph damage} study to investigate their \textit{functional} robustness (accuracy vs. fraction of removed nodes). The \textit{topological} robustness (giant component vs. fraction of removed nodes) is already well-studied in network science. We randomly remove a fixed fraction of nodes, $f$, from a neural network and compute the accuracy achieved by the resulting model on the test dataset. Practically, node removal is implemented using PyTorch's \texttt{Dropout}\footnote{\url{https://pytorch.org/docs/stable/generated/torch.nn.Dropout.html}}, which zeroes some network activations by sampling from i.i.d. Bernoulli distributions. As each batch element is associated with specific random variables, activations produced by different dataset samples are processed by differently pruned neural networks. Therefore, the figure of interest is averaged over the dataset and the 15 generation seeds. In a typical topological analysis, when $f = 0$, the giant components of all tested graphs have the same size (i.e., $N$). We adopt this convention in our experimental setup by replacing test accuracy with \textit{accuracy gain}: $\mathcal{A}(f)$. The metric is defined as the ratio between the accuracy obtained by a pruned network and the accuracy obtained by the original one (i.e., $f = 0$). An accuracy gain $< 1$ indicates a decline in model performance. Consequently, the figure of merit for our analysis is the mean accuracy gain, with the expectation taken over the generation seeds.

\color{black}
\noindent\textbf{The Role of Size and Density.} Size and density are two of the most crucial attributes in a network. In order to explore the influence of these properties on our results, in an additional set of experiments we allowed $N$ and $L$ to vary. This adjustment allowed us to reveal the impact of these two properties on the performance of the models. For clarification, in this context, the size of a network is defined as the number of neurons it contains. Density, on the other hand, refers to the ratio between the actual number of edges and the maximum number of edges an equivalent undirected network with the same $N$ can have: $\rho = \frac{2L}{N(N - 1)}$. As a result, here computational graphs are generated based on a predetermined size/density grid; NNs are then trained and validated using the same synthetic datasets as above, focusing on the four with the highest level of classification difficulty.

\noindent\textbf{Real-World Data.} Synthetic data is valuable for characterizing a model's behavior under varying controllable parameters. However, it is equally pivotal to assess whether the results obtained from the investigated architecture can translate to real-world scenarios. To achieve this goal, we conducted additional experiments using the same ``fair comparison'' framework as outlined in our primary experimental protocol. This framework involved conducting hyperparameter optimization and topology evaluation under consistent conditions, with the same values for $N$ and $L$. We performed these experiments on six datasets sourced from the UCI suite\footnote{\url{https://archive.ics.uci.edu/datasets}}. Specifically, we handpicked the top-6 classification datasets with fewer than 10 numerical attributes, based on their popularity (i.e., number of views). The resulting list, arranged in descending order, comprises: Iris \cite{misc_iris_53}, Glass Identification \cite{misc_glass_identification_42}, Ecoli \cite{misc_ecoli_39}, Rice \cite{misc_rice_(cammeo_and_osmancik)_545}, Breast Cancer Wisconsin \cite{misc_breast_cancer_wisconsin_(original)_15}, and Haberman's Survival \cite{misc_haberman's_survival_43}.
\color{black}

\section{Results}\label{sec:results}

We obtained the presented results by following the experimental protocol outlined in Section \ref{sec:methods} using the specified topologies (i.e., BA, ER, MLP and WS \textcolor{black}{-- further topologies are presented in \ref{sec:hub-based} and \ref{sec:sbm}}) and datasets. We set $\nclasses{} = 3$\footnote{\textcolor{black}{With this particular dataset hyperparameter selection, the neural network's output will consist of a 3-dimensional vector. It's important to note that the input, which represents the 3D coordinates of a sample point on the curve, must also maintain a dimensionality of 3.}} and $\nreps{} \in {3, 6, 9, 12}$; for the \roll{} dataset, $\sigma \in {0.0, 1.0}$, while for the \scurve{}, $\sigma \in {0.0, 0.3}$. The train, validation, and test split sizes were 1350, 675, and 675, respectively. Given that in a 1-hidden layer MLP (\texttt{h1} notation) the number of synaptic connections depends solely on $N$ (i.e., $L = 3 \times H + H \times 3$, with $H = N - 3 - 3$), we chose an MLP with 128 neurons as a reference model and calculated the hyperparameters for the complex networks to achieve graphs with $L = 732$ edges. The additional degree of freedom in the WS generator enabled us to separate the small-world topology into three distinct graph families: \texttt{p.5} ($p = 0.5$), \texttt{p.7} ($p = 0.7$), and \texttt{p.9} ($p = 0.9$). The hyperparameter optimization searched for learning rates in \{0.03, 0.01, 0.003, 0.001\} and batch sizes in \{32, 64\}.

\begin{figure}
\centering
\includegraphics[width=.75\textwidth]{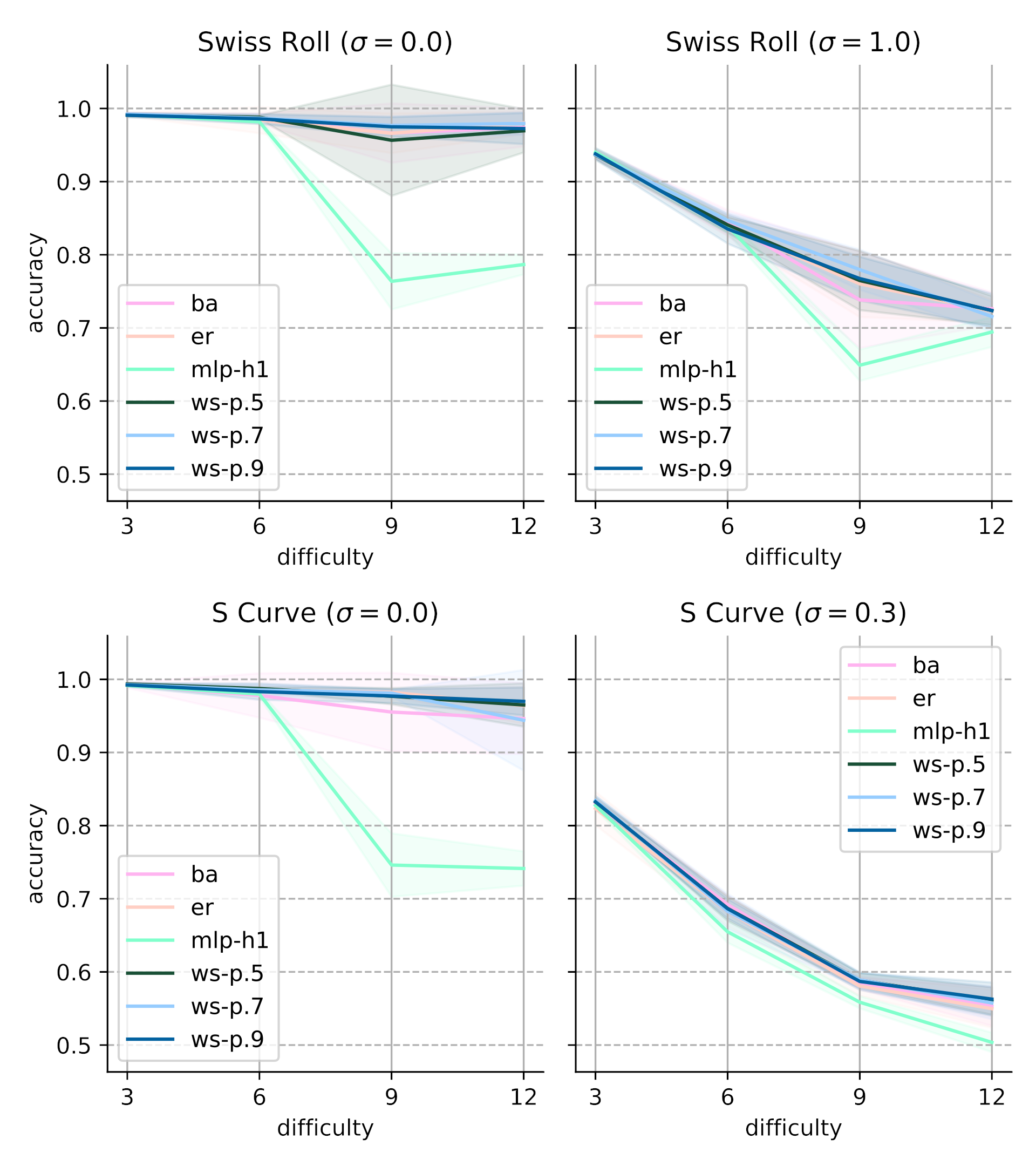}
\caption{Mean test accuracy as a function of the task difficulty. Confidence intervals ($\pm$ standard deviation) are reported as well. Different subplots correspond to different datasets. Each curve denotes the trend of a specific network topology. \label{fig:acc-vs-dif}}
\end{figure}

Figure \ref{fig:acc-vs-dif} displays the mean test accuracy achieved by each group of models as a function of task difficulty. All manifolds, noise levels, and difficulties are represented. Excluding difficulty level 9 in the \roll{} dataset, the accuracy curves exhibit a clear decreasing trend. Specifically, as the difficulty increases, the performance of the MLPs degrades more rapidly than that of complex networks. Confidence intervals, on the other hand, are wider in the high-difficulty plot regions. As expected, noisy tasks were more challenging to learn.

\begin{figure}
\centering
\includegraphics[width=.49\textwidth]{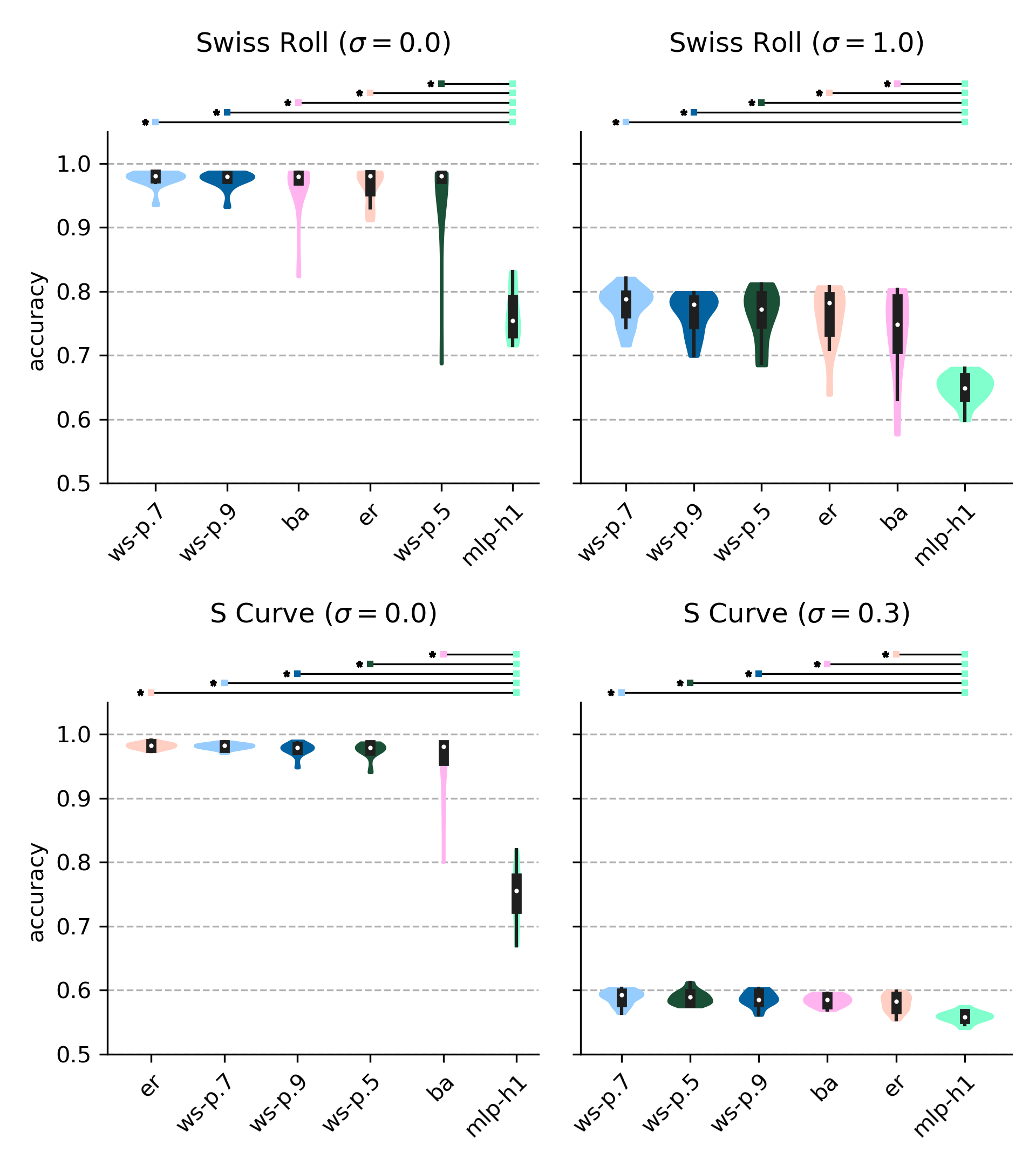}
\includegraphics[width=.49\textwidth]{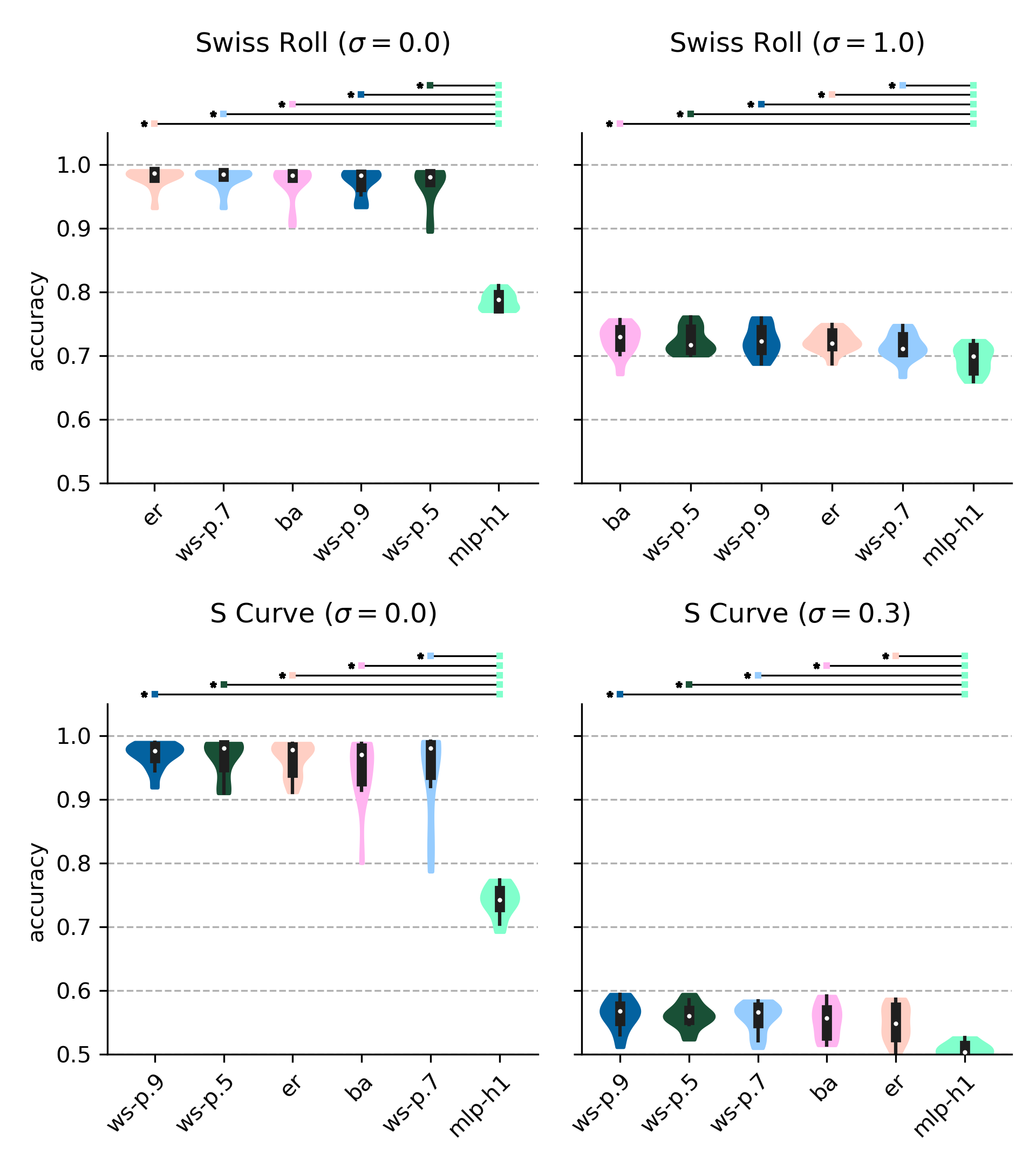}
\caption{\color{black}Accuracy distributions (test split) at the highest difficulty levels. \textbf{Left}: $\text{difficulty} = 9$. \textbf{Right}: $\text{difficulty} = 12$. Each violin corresponds to a specific network topology and is represented by a consistent color across all plots (following the color scheme from Figure \ref{fig:acc-vs-dif}). Statistical annotations appear above the plots, with each segment indicating a significant difference between two medians. Violins in the plots are sorted by mean test accuracy, from left to right in decreasing order. \label{fig:bars}\color{black}}
\end{figure}

In Figure \ref{fig:bars}, the results obtained by the models for the two highest levels of task difficulty are shown in detail. The H-test null hypothesis is rejected for all experiments, and the U-test statistical annotations are displayed. Regardless of the scenario considered, a complex topology consistently holds the top spot in the mean accuracy ranking. MLPs, in contrast, are always the worst-performing models. Moreover, the MLP performance differs significantly from that of the complex networks, in a statistical sense.

\begin{figure}
\centering
\includegraphics[width=.75\textwidth]{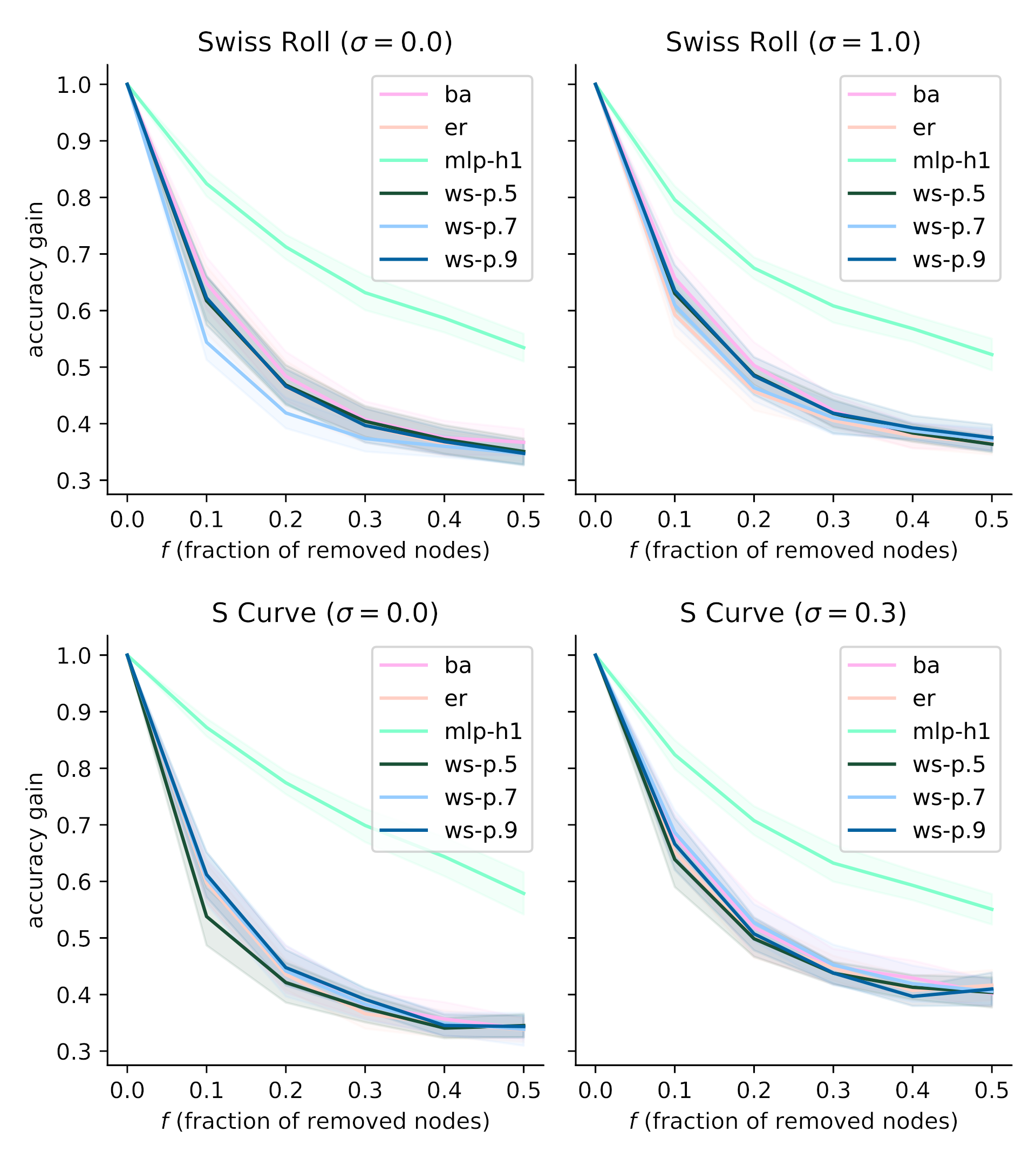}
\caption{Robustness analysis. The horizontal axis reports the fraction of removed nodes (i.e., $f$) while the vertical one the accuracy gain (i.e., $\mathcal{A}(f)$). Each curve refers to a different network topology. Confidence intervals ($\pm$ standard deviation) are reported. \label{fig:robustness}}
\end{figure}

Figure \ref{fig:robustness} presents the results of the robustness analysis. We investigated $f \in \{0.0, 0.1, \dots, 0.5\}$ and removed nodes from the models trained on the datasets characterized by the lowest level of difficulty. On these tasks, indeed, all models behave approximately the same (see Figure \ref{fig:acc-vs-dif}), hinting at a fair comparison. Unsurprisingly, node removal has the same effect on all topologies: the accuracy gain decreases as $f$ increases. MLPs, however, show enhanced robustness to random deletions. Confidence intervals of the complex graph families overlap. It is worth noting that the chance level (i.e., accuracy of $1/3$) could be reached by different accuracy gains depending on the task; the best accuracy under $f = 0$, indeed, varies between the manifold/noise pairs.

\begin{figure}
    \centering
    \includegraphics[width=\textwidth]{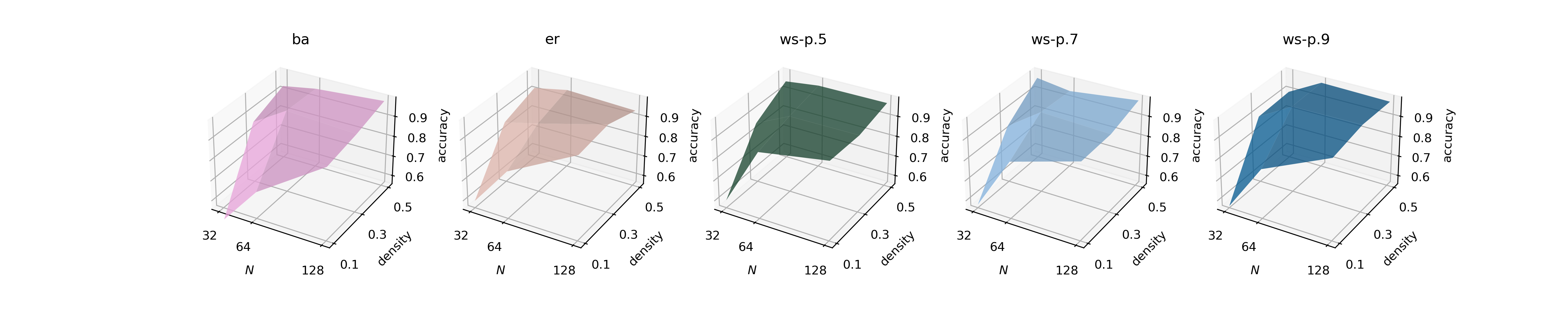}
    \includegraphics[width=\textwidth]{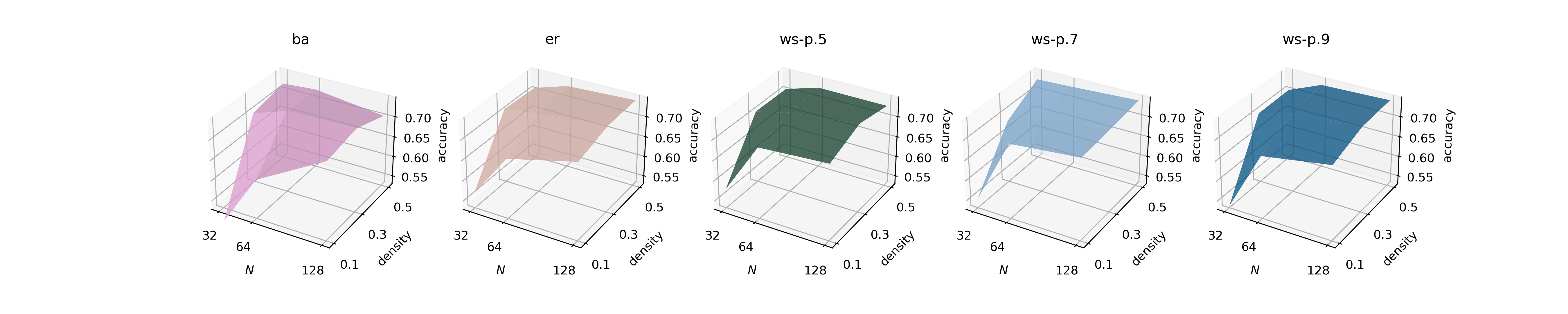}
    \includegraphics[width=\textwidth]{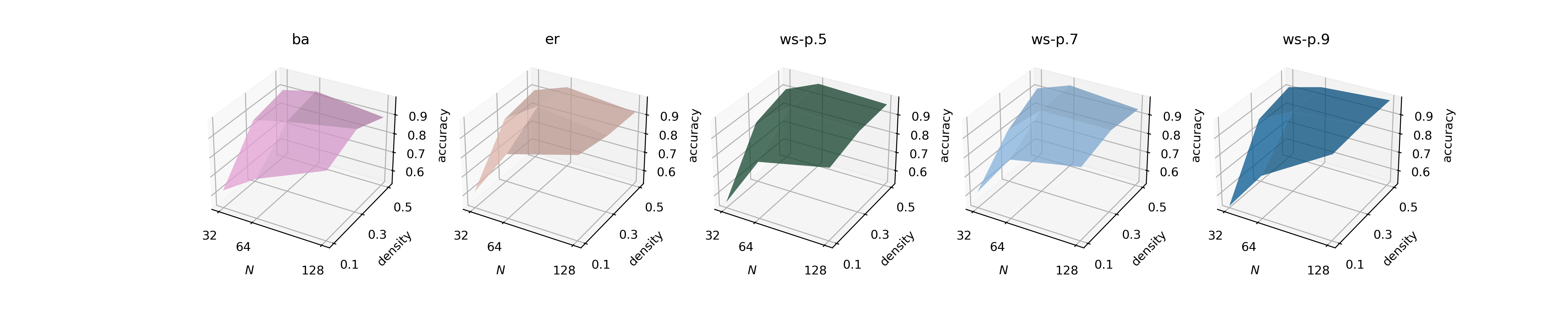}
    \includegraphics[width=\textwidth]{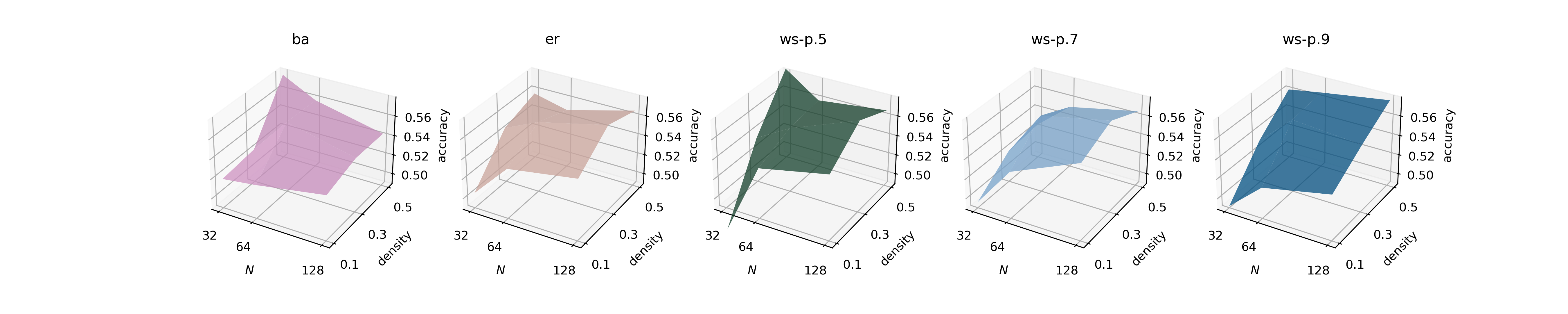}
    \caption{\textcolor{black}{Mean test accuracy as function of size ($N$) and density ($\rho$). From top to bottom: \roll{}, noisy \roll{}, \scurve{} and noisy \scurve{}.}}
    \label{fig:nl}
\end{figure}

\color{black}
The results obtained from the experiments conducted to investigate the influence of size and density on the approximation capabilities of complex NNs are presented in Figure \ref{fig:nl}. It is worth noting that the surfaces displayed, which depict test accuracy points linked to different values for the number of neurons and parameters, exhibit a consistent monotonically increasing pattern for both variables, except for a few outliers. Moreover, in nearly all cases, accuracy saturates as the network size approaches $N = 128$. It is important to clarify that in this series of experiments, each data point represents an average computed over 5 training sessions. Learning rate and batch size were held constant at 0.03 and 64, respectively.

\begin{table}
\centering

\resizebox{\textwidth}{!}{
\begin{tabular}{||l|llllll||}
\hline
Dataset & ba & er & ws-p.5 & ws-p.7 & ws-p.9 & mlp-h1 \\
\hline\hline
breast & \textbf{97.31\%} (±0.27) & 96.82\% (±0.27) & 97.21\% (±0.44) & 97.11\% (±0.42) & 97.21\% (±1.25) & 96.72\% (±0.44) \\
ecoli & 84.95\% (±4.61) & 87.72\% (±1.50) & 81.78\% (±3.33) & 87.52\% (±1.93) & 86.53\% (±2.49) & \textbf{87.92\%} (±0.83) \\
glass & \textbf{69.23\%} (±4.21) & 63.38\% (±5.03) & 68.00\% (±5.48) & 69.23\% (±2.88) & 62.77\% (±9.32) & 64.00\% (±2.06) \\
haberman & 70.45\% (±1.97) & \textbf{72.05\%} (±1.90) & 72.05\% (±1.72) & 70.00\% (±1.90) & 71.14\% (±0.62) & 69.09\% (±0.95) \\
iris & \textbf{93.04\%} (±0.97) & 92.61\% (±1.94) & 92.61\% (±1.19) & 92.61\% (±1.19) & 92.17\% (±1.94) & 92.17\% (±1.19) \\
rice & 92.66\% (±0.25) & 92.47\% (±0.07) & 92.41\% (±0.34) & 92.45\% (±0.18) & 92.47\% (±0.27) & \textbf{92.88\%} (±0.30)\\
\hline
\end{tabular}
}

\caption{\textcolor{black}{Performance of various graph families on the UCI datasets. For each entry, mean test accuracy and standard deviation are reported. On each row, the best result is highlighted in bold.}}
\label{tab:uci}
\end{table}

Finally, Table \ref{tab:uci} presents the performance of the studied models on the UCI real-world data. For each dataset-model pair, we provide the average test accuracy, computed over 5 runs, along with the associated standard deviation. It is worth noting that these experiments were conducted following the ``fair comparison'' setup, wherein models trained on the same dataset possess an equal number of neurons (128) and parameters. Across different datasets, the number of parameters varies as it is determined by the number of hidden neurons, which is set as $N$ - \textit{the number of features} - \textit{the number of classes}. The number of edges, L, follows the equations introduced at the beginning of this section. Hyperparameters were optimized as previously described for the experiments in line with the main experimental protocol. In 4 out of 6 datasets, complex NNs outperformed MLPs, whereas in the remaining cases, MLPs exhibited comparable performance to complex NNs.
\color{black}

\section{Discussion}\label{sec:discussion}

\textcolor{black}{In this paper, the most significant finding is the performance, in terms of accuracy, attained by the architectures built on complex topologies both in high-difficulty scenarios using synthetic data and in classification problems defined within the UCI dataset suite. In this context, and in light of the statistical tests carried out, the complex models prove to be a solid alternative to MLPs.} 

Formally justifying the observed phenomenon is challenging. Fortunately, in 2017, Poggio et al. discussed two theorems \cite{poggio} that guided our explanation. According to the first theorem\footnote{We invite the reader to consult ref. \cite{poggio} for a complete formulation of the theorems.}, a shallow network (e.g., an MLP \texttt{h1}) equipped with infinitely differentiable activation functions requires $N = \mathcal{O}(\epsilon^{-n})$ units to approximate a continuous function $f$ of $n$ variables\footnote{Depending on the context, we use the same $f$ notation for both the fraction of removed nodes and the function to be approximated.} with an approximation error of at most $\epsilon > 0$. This exponential dependency is technically called the \textit{curse of dimensionality}. On the other hand, the second theorem states that if $f$ is compositional and the network presents its same architecture, we can escape the ``curse''. It is important to remember that a compositional function is defined as a composition of ``local'' constituent functions, $h \in \mathcal{H}$ (e.g., $f(x_1, x_2, x_3) = h_2(h_1(x_1, x_2), x_3)$, where $x_1,\ x_2,\ x_3$ are the input variables and $h_1,\ h_2$ the constituent functions). In other words, the structure of a compositional function can be represented by a DAG. In this approximation scenario, the required number of units depends on $N = \mathcal{O}(\sum_{h} \epsilon^{-n_h})$, where $n_h$ is the input dimensionality of function $h$. If $\max_{h} n_h = d$, then $\sum_{h} \epsilon^{-n_h} \le \sum_{h} \epsilon^{-d} = \abs{{\mathcal{H}}}\epsilon^{-d}$.

The primary advantage of complex networks is their potential to avoid the curse of dimensionality when relevant graphs for the function to be learned are present. Under the assumption that the function linking the \roll{} and \scurve{} points to the ground truth labels is compositional (intuitively, in non-noisy datasets, each class is a union of various segments), we conjecture that our complex NNs can exploit this compositionality. In the high-difficulty regime, the necessary network size for MLP \texttt{h1} to achieve the same accuracy as complex models likely exceeds the size set for experiments. While one could argue that the datasets employed were compositionally sparse by chance, according to \cite{poggio_it_nodate}, all \textit{efficiently computable functions} must be \textit{compositionally sparse} (i.e., their constituent functions have ``small'' $d$). Performance differences on noisy datasets are less noticeable, possibly due to the minimal overlap between the functions to be approximated and the studied topologies. Notably, our setup does not precisely match the theorem formulations in \cite{poggio} (e.g., SELUs are not infinitely differentiable), but Poggio et al. argue that the hypotheses can likely be relaxed. No statistically significant differences emerged between the complex graph families from the results of Section \ref{sec:results}. Various explanations exist for this outcome: all tested topologies could be complex enough to include relevant subgraphs of the target $f$ functions; the random DAG conversion heuristic might have perturbed hidden topological properties of the original undirected networks; or the degree distribution of a network may not be the most relevant topological feature in a model's approximation capabilities. \textcolor{black}{Additionally, the results in Table \ref{tab:uci} demonstrate that the optimal topology of a NN is task-specific. This suggests the need for future advancements in the state of the art of neural architecture search, potentially formulating the creation of computational graphs through data-driven approaches.}

The higher accuracy in complex networks, however, comes with trade-offs. Although the methodology in \cite{https://doi.org/10.48550/arxiv.2209.02037} improves the scalability of complex NNs and enables experimentation with arbitrary DAGs, it is important to note that 1-hidden layer MLPs typically have faster forward pass computation. In these models, the forward pass requires only two matrix multiplications, whereas, in NNs built using 4Ward, the number of operations depends on the DAG \textit{height}. \textcolor{black}{However, we believe that, in the future, the computational efficiency of tools like 4Ward will be enhanced through the integration of ML frameworks optimized for sparse tensor processing \cite{pmlr-v202-nikdan23a} and specialized hardware \cite{DBLP:journals/corr/abs-2212-02872}.} Moreover, the analyses in Figure \ref{fig:robustness} demonstrate MLPs' superiority in a graph damage scenario. We speculate that the hidden units in an MLP \texttt{h1} contribute equally to the approximation of the target function. In contrast, the ability of complex networks to exploit the compositionality of the function to be learned might lead to high specialization of some hidden units.

\section{Conclusions}\label{sec:conclusion}

Our study provides valuable insights into the influence of network topology on the approximation capabilities of artificial neural networks (ANNs). Our novel methodology for constructing complex ANNs based on various topologies has enabled a systematic exploration of the impact of network structure on model performance. The experiments conducted on synthetic datasets demonstrate the potential advantages of complex topologies in high-difficulty regimes when compared to traditional MLPs.

While complex networks exhibit improved performance, this comes at the cost of increased computational requirements and reduced robustness to graph damage. Our investigation of the relationship between topological attributes and model performance (Appendix \ref{sec:attributes}) reveals a complex interplay that cannot be explained by any single attribute. This finding highlights the need for further research to better understand the interactions among multiple topological attributes and their impact on ANN performance.

As a result of this study, researchers and practitioners can consider the potential benefits and limitations of complex topologies when designing ANNs for various tasks. Moreover, our work provides a foundation for future research focused on identifying optimal topological features, understanding the impact of multiple attributes, and developing new methodologies for constructing more efficient and robust ANN architectures. By further exploring the role of network topology in ANNs, we can unlock new possibilities for improving the performance and adaptability of these models across diverse applications.

\section*{Acknowledgements}

This work is supported and funded by: \#NEXTGENERATIONEU (NGEU); the Ministry of University and Research (MUR); the National Recovery and Resilience Plan (NRRP); project MNESYS (PE0000006, to NT) - \textit{A Multiscale integrated approach to the study of the nervous system in health and disease} (DN. 1553 11.10.2022); the MUR-PNRR M4C2I1.3 PE6 project PE00000019 Heal Italia (to NT); the NATIONAL CENTRE FOR HPC, BIG DATA AND QUANTUM COMPUTING, within the spoke \textit{``Multiscale Modeling and Engineering Applications''} (to NT); the European Innovation Council (Project CROSSBRAIN - Grant Agreement 101070908, Project BRAINSTORM - Grant Agreement 101099355); the Horizon 2020 research and innovation Programme (Project EXPERIENCE - Grant Agreement 101017727). Tommaso Boccato is a PhD student enrolled in the National PhD in Artificial Intelligence, XXXVII cycle, course on Health and Life Sciences, organized by Università Campus Bio-Medico di Roma.

\appendix

\section{Graph Attributes}\label{sec:attributes}

\begin{figure}[h]
\centering
\includegraphics[width=.7\textwidth]{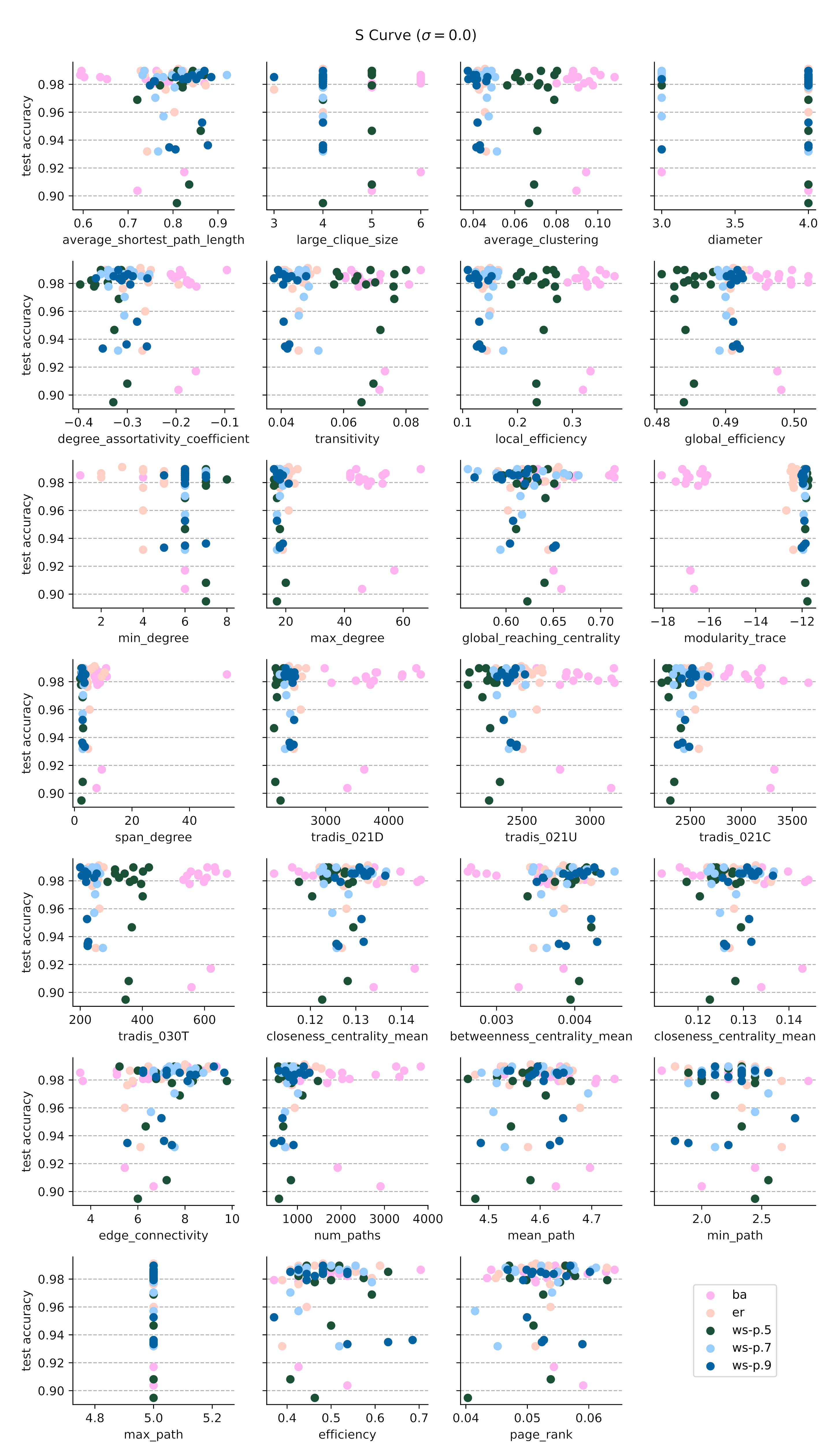}
\caption{Correlation plots (accuracy vs. attribute) computed on the \roll{} dataset ($\nreps{} = 12,\ \sigma = 0.0$). Each network topology is denoted with a different color, which can be found in the legend (last subplot). \label{fig:attributes-roll}}
\end{figure}

\begin{figure}[h]
\centering
\includegraphics[width=.7\textwidth]{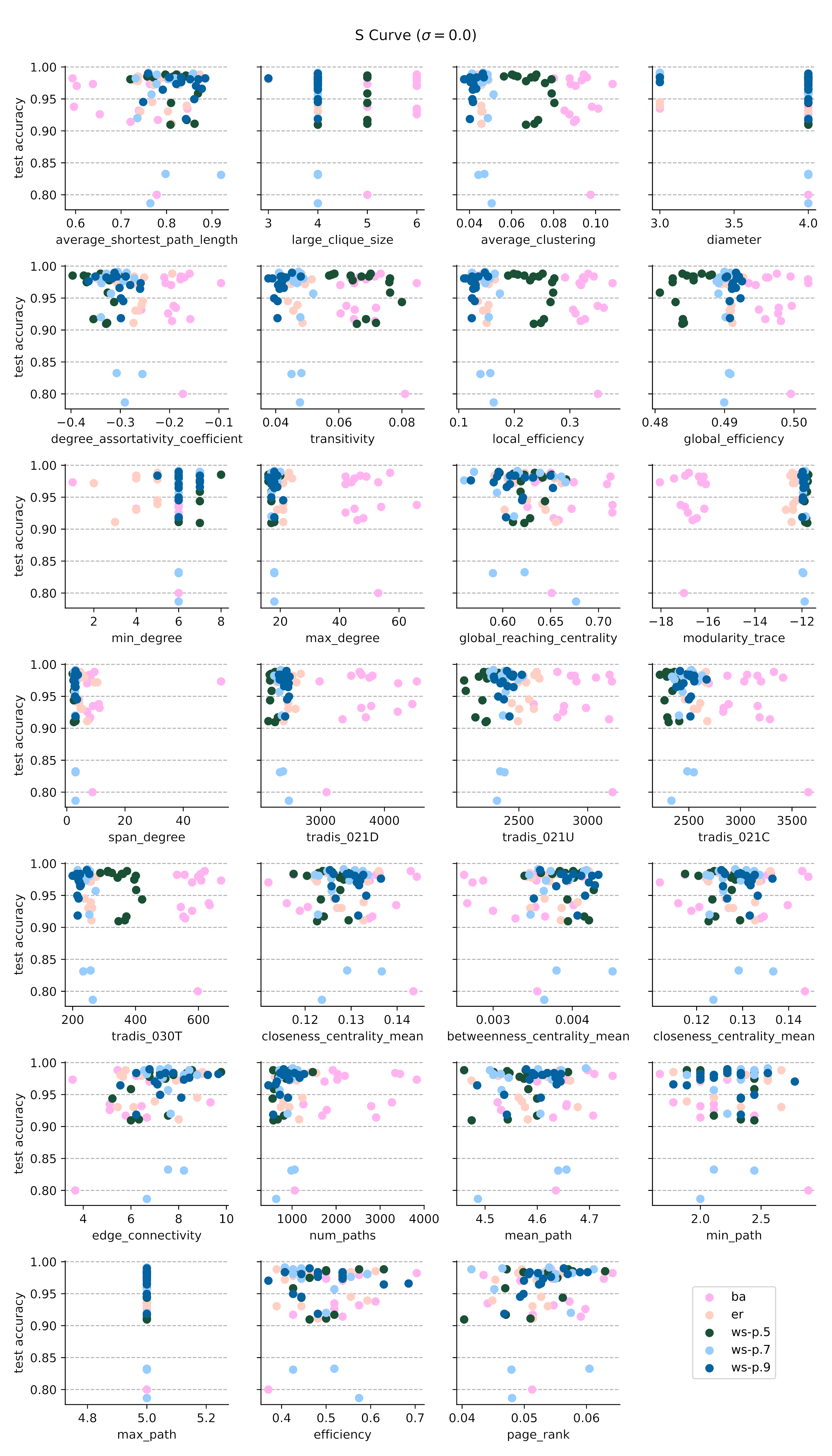}
\caption{Correlation plots (accuracy vs. attribute) computed on the \scurve{} dataset ($\nreps{} = 12,\ \sigma = 0.0$). Each network topology is denoted with a different color, which can be found in the legend (last subplot). \label{fig:attributes-s}}
\end{figure}

We delved deeper into the role of network topology in the models' approximation capabilities by calculating a total of 27 topological graph attributes for each trained neural network. Our aim was to ascertain if any specific attributes could account for the models' performance. The correlation plots that display the relationship between test accuracy and graph attribute can be found in Figures \ref{fig:attributes-roll} and \ref{fig:attributes-s}.

To compute these metrics, we employed the NetworkX library \cite{SciPyProceedings_11} when feasible, and devised custom implementations for the remaining attributes. We conducted the experiments using noise-free datasets with the highest difficulty levels.

After analyzing the experimental data, no evident relationship emerged between the attributes and the models' performance. In other words, when examined individually, none of the attributes could account for the achieved accuracies. This outcome implies that the impact of network topology on the approximation capabilities of the models might be more intricate than a straightforward correlation with any single topological attribute. Further investigation is necessary to explore the potential interplay among multiple attributes and their influence on the models' performance.

\color{black}
\section{Hub-Based Topological Orderings}\label{sec:hub-based}

When creating an undirected BA graph, a significant number of hubs emerge alongside numerous low-degree nodes. This raises the question of what happens when, during the DAG conversion, nodes are  arranged based on their degree. To investigate this, we have chosen to focus on three key topological orderings: sorting nodes by degree in descending order, sorting nodes by degree in ascending order, and sorting nodes in descending order but starting from the center of the node sequence and extending towards the edges. In other words, the input features can be assigned to the largest hubs, the output logits can be extracted from the largest hubs, or high-degree nodes can be positioned in the middle of the information flow.

\begin{figure}
\centering
\includegraphics[width=.75\textwidth]{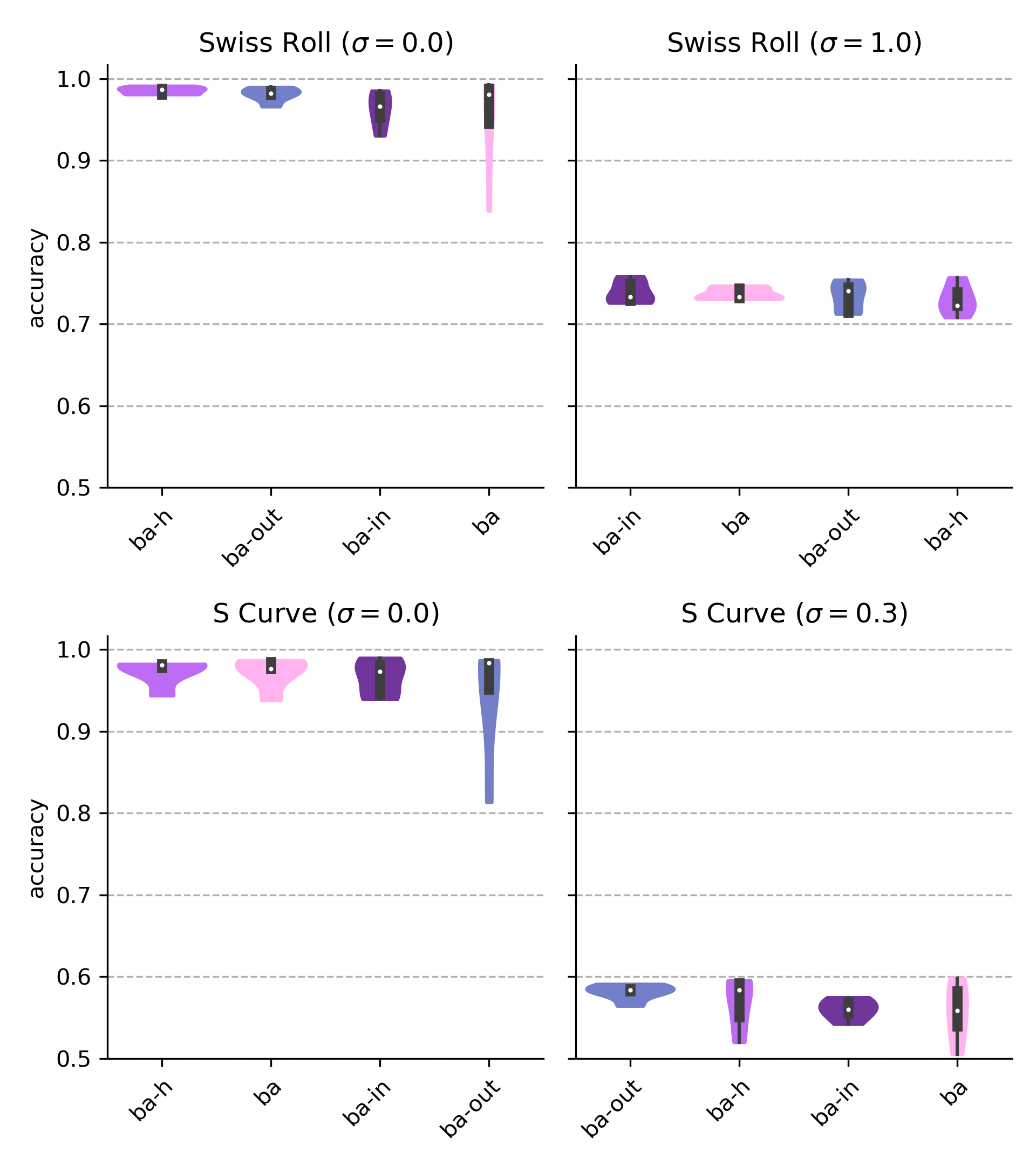}
\caption{\textcolor{black}{Accuracy distributions (test split) for BA NNs generated through different degree-aware topological sortings. Violins in the plots are sorted by mean test accuracy (from left to right).}\label{fig:hub-based}}
\end{figure}

For each new topological ordering, we trained 5 models, each with 128 neurons and 732 parameters, on the \roll{} and 
\scurve{} datasets characterized by the highest level of difficulty. The hyperparameter search and training procedures followed the same methodology as in all other experiments in this paper. The accuracy distributions resulting from evaluating these models on the respective test splits are depicted in Figure \ref{fig:hub-based}. Here the group labeled as ``ba'' represents computational graphs that served as a baseline. These baseline graphs were obtained by converting them through a standard random topological sorting. Regrettably, our experiments did not reveal any statistically significant differences between the various model families. This outcome allows for two distinct interpretations: either the placement of hubs within the topological orderings, which determine the direction of connections, does not significantly impact the models' approximation capabilities, or the NNs employed do not possess a sufficient number of nodes to render hubs a critical factor during the training process. Scaling these experiments in the future would potentially address this ambiguity.

\section{The Stochastic Block Model}\label{sec:sbm}

The ER, BA, and WS models are among the most well-known tools used for generating undirected networks that exhibit non-trivial degree distributions; however, this is not an exhaustive list. For instance, the Stochastic Block Model (SBM) is another widely recognized generative model for random graphs. It establishes connections between nodes based on their membership in specific communities. The generative algorithm for the SBM requires two key parameters: a partition of the vertex set and a symmetric matrix $P$, which contains edge probabilities. Each element $P_{ij}$ in the matrix defines the probability of nodes from community $i$ connecting with nodes from community $j$. When all entries in matrix $P$ are constant, the model reverts to the ER one. Conversely, if both the diagonal and off-diagonal entries are equal, we refer to it as the \textit{planted partition model}. In this case, if we denote the intra-community probability as $p$ and the inter-community probability as $q$, the model is categorized as \textit{assortative} when $p > q$ and disassortative when $p < q$.

\begin{figure}
\centering
\includegraphics[width=.75\textwidth]{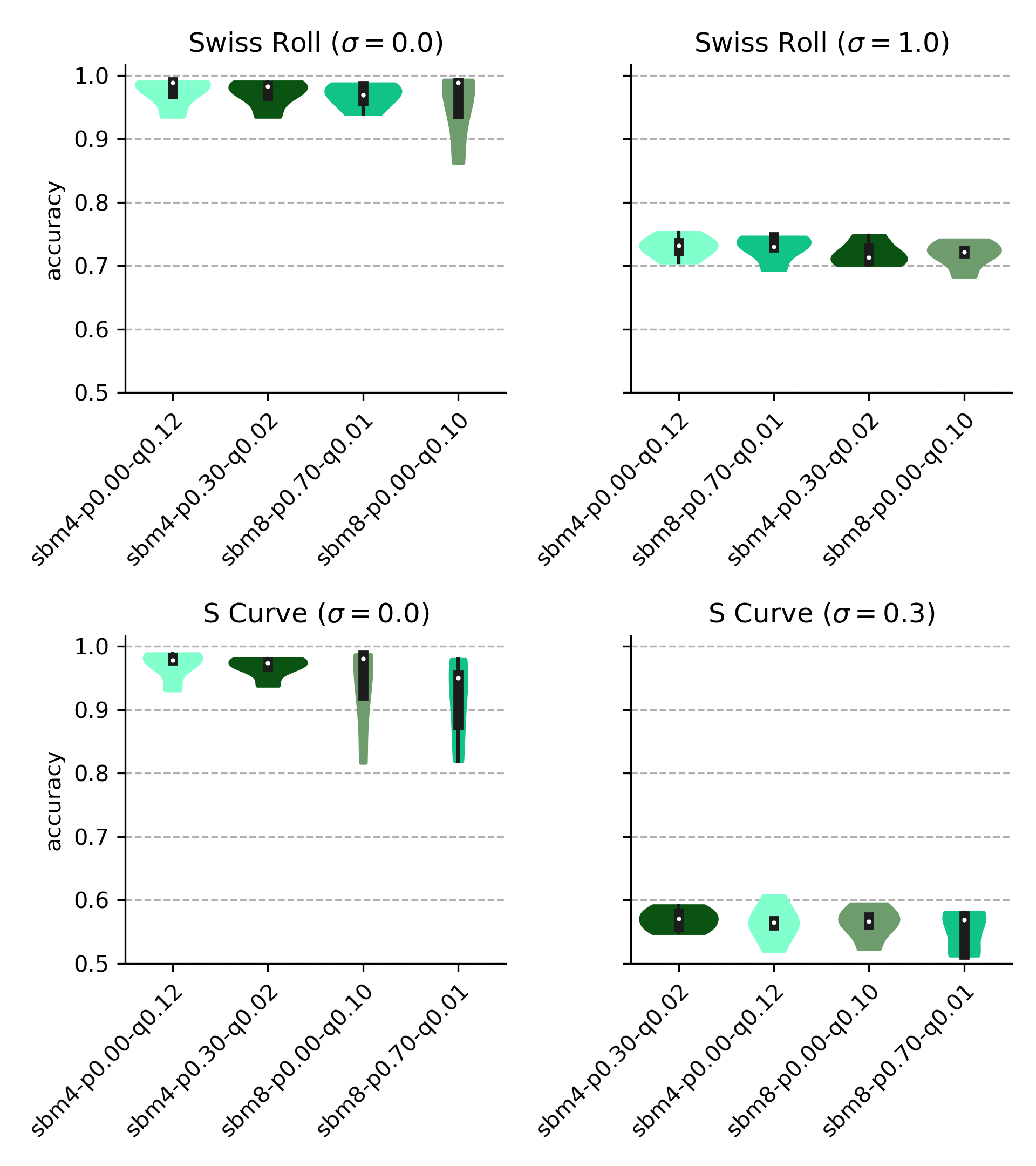}
\caption{\textcolor{black}{Accuracy distributions (test split) for various SBM NNs. Model names follow the following coding: \texttt{sbm$x$-p$y$-q$z$} where $x$, $y$ and $z$ represent the number of communities, the intra-community and the inter-community probability, respectively. Violins in the plots are sorted by mean test accuracy (from left to right).}\label{fig:sbm}}
\end{figure}

Due to the numerous parameters involved in the generation process, we conducted an experiment focused on the planted partition model. The objective was to investigate whether this topology could effectively serve as the basis for a neural network and to evaluate the impact of partition size, intra- and inter-community probabilities on the overall performance of the generated architectures. We set two different partition sizes, namely 4 and 8, with $N=128$ and $\mathbb{E}[L] = 732$. For each number of communities, we examined two distinct scenarios: the assortative model characterized by the highest possible $p$, and the disassortative one with the highest possible $q$, in an effort to explore the as diverse configurations as possible. In the first scenario, we can observe highly dense clusters with sparse inter-cluster connections, while in the second scenario, sparse bipartite graphs within a fully-connected meta-graph. For each configuration, we trained 5 models after conducting a preliminary hyperparameter search. Results are reported in Figure \ref{fig:sbm}. While optization have conferged in all experiments, differences in accuracy distributions among the models are not statistically significant. Nevertheless, it is worth noting that the 4-community graphs consistently exhibit the highest average test accuracy across most of the tested tasks.

\color{black}

\section{Glossary}

\begin{itemize}
    \item Artificial Neural Network (ANN)
    \item Barabási-Albert (BA)
    \item Directed Acyclic Graph (DAG)
    \item Erdős-Rényi (ER)
    \item Lottery Ticket Hypothesis (LTH)
    \item Multilayer Perceptron (MLP)
    \item Neural Network (NN)
    \item Root Mean Square Error (RMSE)
    \item Watts-Strogatz (WS)
\end{itemize}

\clearpage
\bibliographystyle{elsarticle-num}
\bibliography{refs}





\end{document}